\documentclass[preprint,12pt]{elsarticle}
\makeatletter
\def\ps@pprintTitle{%
 \let\@oddhead\@empty
 \let\@evenhead\@empty
 \def\@oddfoot{\reset@font\hfil\thepage\hfil}%
 \let\@evenfoot\@oddfoot}
\makeatother

\usepackage{subcaption, graphicx}

\usepackage{amssymb}
\usepackage{amsmath}
\usepackage{booktabs}
\usepackage{xcolor, soul, float}

\begin{document}

\begin{frontmatter}



\title{ESCAPE: Energy-based Selective Adaptive Correction for Out-of-distribution 3D Human Pose Estimation}


\author[1]{Luke Bidulka}
\author[1]{Mohsen Gholami}
\author[1]{Jiannan Zheng}
\author[2]{Martin J. McKeown}
\author[1]{Z. Jane Wang}

\affiliation[1]{organization={Department of Electrical and Computer Engineering, University of British Columbia},
            city={Vancouver},
            state={BC},
            country={Canada}
            }
\affiliation[2]{organization={Department of Medicine, University of British Columbia},
            city={Vancouver},
            state={BC},
            country={Canada}
            }

\begin{abstract}

Despite recent advances in human pose estimation (HPE), poor generalization to out-of-distribution (OOD) data remains a difficult problem. While previous works have proposed Test-Time Adaptation (TTA) to bridge the train-test domain gap by refining network parameters at inference, the absence of ground-truth annotations makes it highly challenging and existing methods typically increase inference times by one or more orders of magnitude. We observe that 1) not every test time sample is OOD, and 2) HPE errors are significantly larger on distal keypoints (wrist, ankle). To this end, we propose ESCAPE: a lightweight correction and selective adaptation framework which applies a fast, forward-pass correction on most data while reserving costly TTA for OOD data. The free energy function is introduced to separate OOD samples from incoming data and a correction network is trained to estimate the errors of pretrained backbone HPE predictions on the distal keypoints. For OOD samples, we propose a novel self-consistency adaptation loss to update the correction network by leveraging the constraining relationship between distal keypoints and proximal keypoints (shoulders, hips), via a second ``reverse" network. ESCAPE improves the distal MPJPE of five popular HPE models by up to 7\% on unseen data, achieves state-of-the-art results on two popular HPE benchmarks, and is significantly faster than existing adaptation methods.
\end{abstract}

\begin{keyword}
3D Human Pose \sep Test-Time Adaptation \sep Energy Function \sep Out-of-distribution Detection



\end{keyword}

\end{frontmatter}

\section{Introduction}
\label{sec:intro}
3D Human Pose Estimation (HPE) from single-view images is a fundamental task in computer vision with applications in healthcare \cite{GHOLAMI2023102871}; \cite{yu2023pa}, VR/AR \cite{Vid2Player}, human-robot-interaction \cite{zhu_2020_eccv_nba}, etc.
Recent deep-learning-based HPE methods have achieved remarkable performance \cite{spin,hybrik,PARE,cliff} on public benchmarks.
Due to the immense difficulty of labeling 3D human pose data, most HPE methods are trained using public datasets which do not accurately represent the wide variety inherent in in-the-wild poses.
Therefore, the pre-trained models significantly under-perform when applied to out-of-distributions (OOD) samples \cite{adaptpose,boa}. The performance of HPE models specifically degrades in estimating the 3D distal keypoints (e.g. wrist, ankle) \cite{weng2022domain,CycleAdapt}.

\begin{figure} [t]
    \centering
    \includegraphics[scale=0.3]{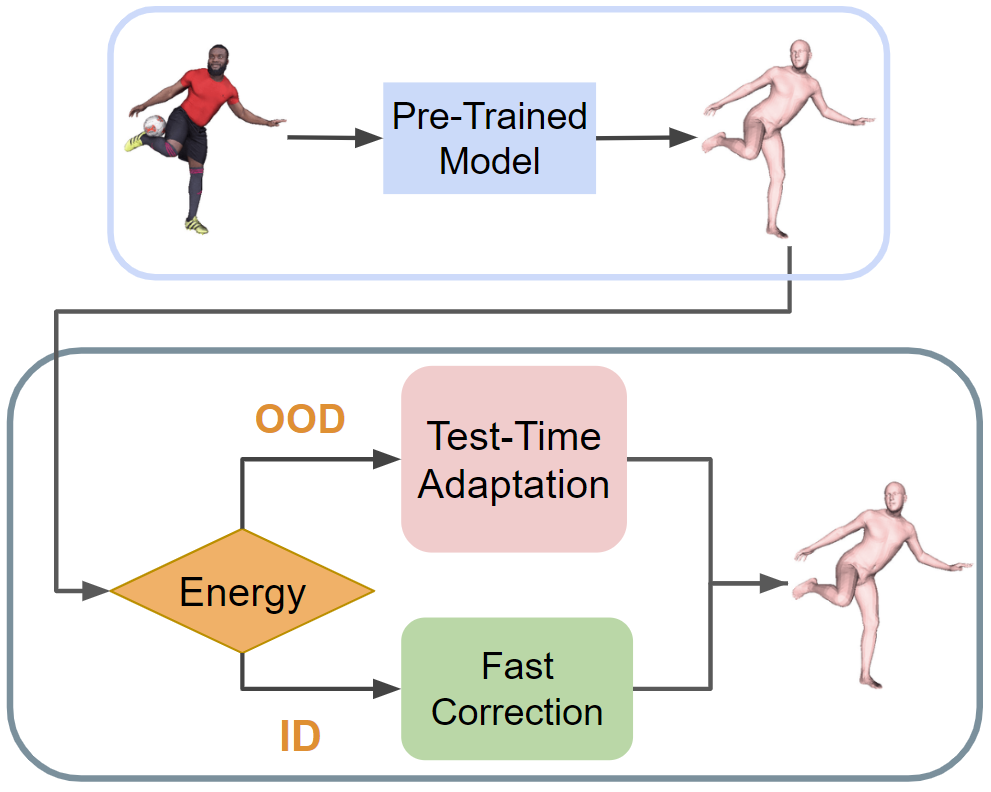}
    \caption{High-level illustration of ESCAPE, which improves backbone model predictions by seperating harder, out-of-distribution samples and easier, in-distribution samples via an energy function (Sec. \ref{sec:energy}) and applying intensive test-time adaptation (Sec. \ref{sec:TTA}) or fast forward-pass correction (Sec. \ref{sec:cNet}) respectively.}
    
    \label{fig:framework_high-level}
\end{figure}

Recently, test-time adaptation (TTA) methods have been shown to help bridge the train-test domain gap and improve prediction performance on OOD data by fine tuning models on test data \cite{boa,weng2022domain,CycleAdapt, ISO}. 
Most of the prior works leverage the ground-truth 2D poses of the test sample to optimize the pre-trained models via 2D projection. However, ground-truth 2D poses are not available in real case scenarios and replacing them with estimated 2D keypoints significantly degrades the accuracy of prior works \cite{CycleAdapt}. 
Moreover, because existing SOTA test-time adaptation methods indiscriminately optimize network parameters for every test sample, they massively increase inference times and remain impractical (e.g. BOA \cite{boa} and DynBOA \cite{dynboa} increase inference times by over 50x). 

To address the above issues, we propose \emph{ESCAPE}: Energy-based Selective-adaptive Correction for out-of-distribution 3d humAn Pose Estimation. A fast, selective test time adaptation method which only performs costly test time adaptation on OOD samples, optimizes an external lightweight correction network instead of the pre-trained model, and requires no 2D or 3D annotations whatsoever. Fig. \ref{fig:framework_high-level} shows the high-level overview of \emph{ESCAPE}. 

\emph{ESCAPE} uses a \textit{Free Energy Function} to classify the pre-trained model  outputs as corresponding to OOD or in-distribution (ID) inputs for the pre-trained model.
This energy function is a label-free OOD detection that has shown to be fast and robust \cite{energy}. 
We then apply an intensive test-time adaptation for OOD samples and a fast forward-pass correction for ID samples.

Our proposed test-time adaptation method relies on two observations. 1) pre-trained models estimate the proximal keypoints (e.g. hip and shoulder) with higher confidence compared with distal keypoints (e.g. wrist and ankle).
That is, the majority of 3D keypoint estimation errors are in the distal keypoints because they are often poorly captured in a given frame (blurry, occluded, etc) and are highly unconstrained. Conversely, the proximal keypoints are generally predicted well by virtue of being relatively constrained usually quite visible.
We hypothesize that with proper considerations, we can leverage the well-estimated proximal keypoints to help correct the estimated distal keypoints. 
2) due to biomechanical human joint constraints, there exists a correlation between distal and proximal keypoints. Specifically, given a set of distal keypoints there is a small space of possible corresponding proximal keypoints.
We qualitatively demonstrate the above two observations in the experiments.

\begin{figure*} [t]
    \centering
    \includegraphics[width=\textwidth]{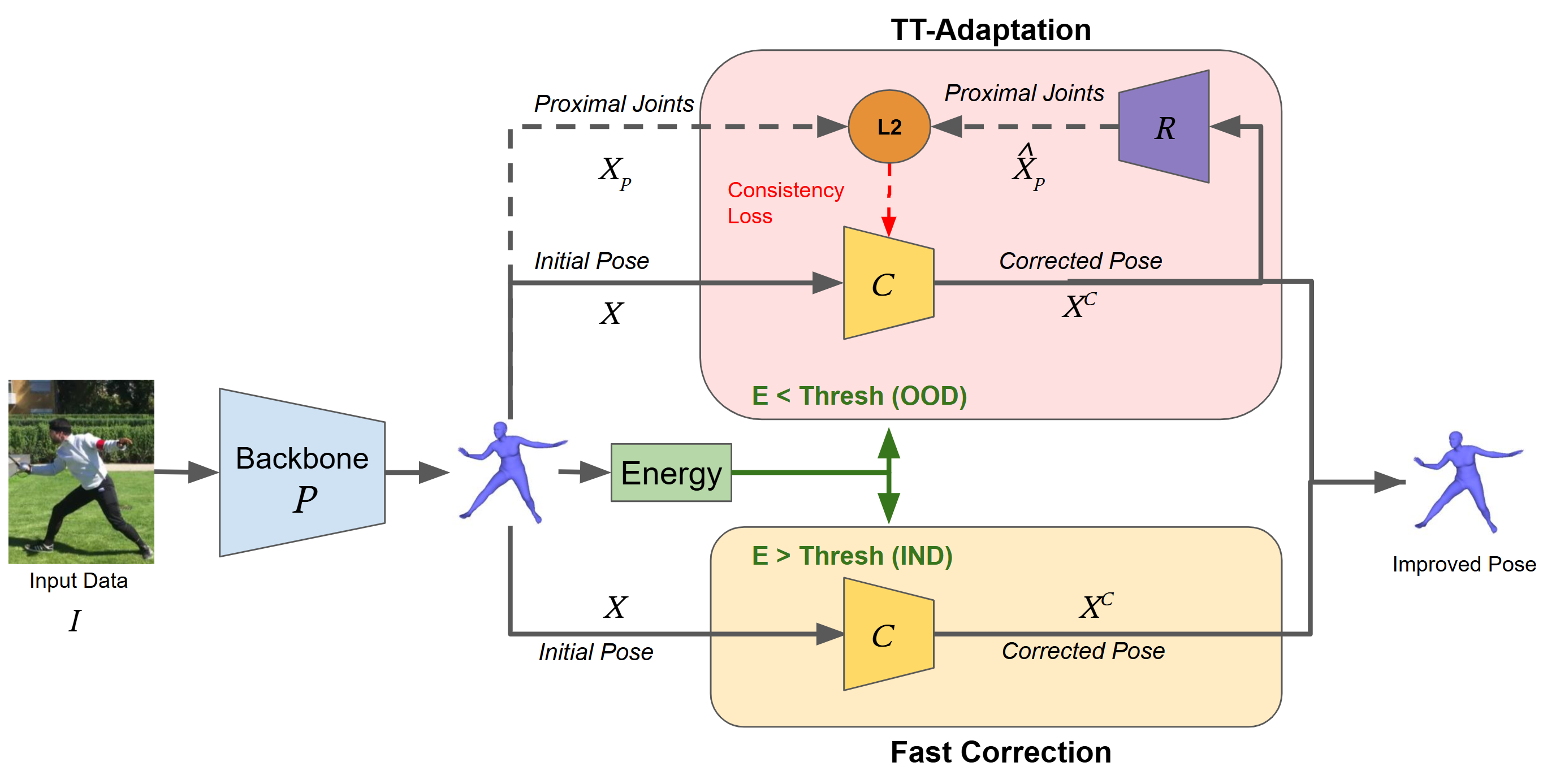}
    \caption{Detailed overview of ESCAPE, the proposed selective adaptation and correction framework for 3D human pose correction. Given an input sample (I), a pre-trained backbone human pose estimator(P) predicts an initial pose (X) and the input sample is classified as in-distribution (ID) or out-of-distribution (OOD) for the backbone model by comparing the its energy score to a predetermined threshold. If the sample was an easier ID sample, only the fast forward pass distal joint correction of $\mathcal{C}$ is applied to produce the final improved pose. If instead the sample was a harder OOD sample, intensive test-time adaptation (TT-Adaptation) is used to fine-tune $\mathcal{C}$ to the current sample and the distal correction from the adapted $\mathcal{C}$ subsequently produces the final corrected pose.}
    \label{fig:framework}
\end{figure*}

Following these observations, the \textit{Fast Correction} (Fig. \ref{fig:framework_high-level}) simply estimates the errors of distal keypoints given the initial estimated pose from the pre-trained model. We show that for ID samples, we can learn a correction network (CNet) to estimate the distal keypoint errors and thereby effectively correct the poses.
For OOD samples, we propose a novel \textit{Test-time Adaptation} utilizing the learned CNet in tandem with a reverse correction network (RCNet) which estimates the errors of proximal keypoints given the CNet corrected pose. We pre-train RCNet with predictions from CNet and use it to optimize CNet at test time via a \textit{Self-Consistency Loss} comparing the RCNet-corrected proximal joints to the intially estimated proximal joints from the pre-trained backbone estimator (Fig. \ref{fig:framework}).

In contrast to prior work, our test-time correction and adaptation method does not optimize the pre-trained model and so can be added downstream of any pre-trained pose estimator to boost performance on both ID and OOD samples. 
Further, because the proposed method optimizes an external correction network instead of the pre-trained model, and only does so on OOD samples, it maintains practical and comparable model inference times and is many times faster than other methods.

Our \textbf{contributions} are as follows:
\begin{itemize}

    \item We propose a novel approach for test-time adaptation methods in which expensive adaptation is reserved for difficult OOD samples and otherwise only a fast feed-forward correction is applied, and demonstrate that an Energy function can be effectively used to perform the OOD sample selection.
    \item We propose a novel and lightweight method (CNet) for correcting pre-trained backbone pose estimator predictions and a novel test-time adaptation method for improving CNet via a learned self-consistency loss using only the backbone estimations, avoiding direct optimization of backbone model parameters.
    \item We attain state-of-the-art results on three popular 3D pose estimation benchmark datasets, atop several pre-trained pose estimation models.
    \item We provide an extensive set of experiments and ablation studies, demonstrating the benefits of our Energy based sample selection for existing TTA methods and verifying the robustness of our proposed method across environmental conditions.
\end{itemize}
\section{Related Work}
\label{sec:related_work}

\textbf{3D Human Pose Estimation}
There are two categories of 3D human pose estimation models: parametric and non-parametric methods. The parametric methods estimate human joint angles for a parametric human mesh model (SMPL) and use that to obtain human mesh and pose \cite{PARE,cliff,spin}. Non-parametric methods directly estimate 3D positions of human keypoints given images \cite{videopose3d,Transformer,MHFormer}. Some of the non-parametric methods divide the problem into two stage, first estimating 2D poses from the image and then estimating 3D poses from 2D poses. Our test-time adaptation method is applicable to all of the above-mentioned categories since it merely requires estimated 3D keypoints output by the pre-trained model for its optimization. 

\textbf{Training-Time Adaptation.}
Recently, the generalizability of HPE models has gained attention. 
Prior work has proposed various solutions to address it at training time via synthetic data generation or modifying the architecture of the network. \cite{Li_2020_CVPR}, \cite{gong2021poseaug}, and \cite{adaptpose}, proposed synthetic data generation to mitigate the domain gap between train and test samples. 
They generate 2D-3D novel poses from novel viewpoints to cover unseen poses and viewpoints in the test dataset. 
Such methods are specifically effective when the ground truth 2D poses are available at the inference time and often struggle when estimated 2D keypoints are not accurate \cite{adaptpose}. 
\cite{SRNet} addresses the poor generalizability of HPE models by splitting the human body into body parts. 
They argue that the splitting procedure helps in estimating poses that are not available in the training dataset since sub-body poses have been seen in the training data. 
\cite{wang2020predicting} proposes estimating camera viewpoint along with 3D poses to address the generalizability of HPE models.
All of the above-mentioned methods are training-time adaptation methods and are unable to dynamically adapt the HPE models.

\textbf{Test-Time Adaptation.} In this section we discuss the prior works that optimize the pre-trained model at the inference stage to address the generalizability problem. 
ISO \cite{ISO} proposes using geometric cycle consistency loss to update the pre-trained HPE model at the inference stage. The proposed loss is inspired by an unsupervised loss that learns 3D poses from 2D poses without any 3D supervision \cite{chen2019unsupervised}. 
BOA \cite{boa} proposes a bilevel optimization method that relies on 2D keypoints and temporal information of the test samples. BOA optimizes the pre-trained models on each single frame to minimize the 2D projection loss. 
DynBOA further improves BOA by retrieving examples from the train data that can specifically help in inference on a test image. DAPA \cite{weng2022domain} proposes a domain-adaptive pose augmentation method that augments the estimated pose given a target image.
Then, the body regression network is finetuned on both the real target images and their augmented counterparts.
The above methods highly depend on the error of test sample 2D keypoints. When 2D keypoints are noisy the performance drops significantly \cite{CycleAdapt}. 
CycleAdapt \cite{CycleAdapt} mitigates the above problem by proposing a motion-denoising network that corrects the 3D estimated motions at the test time and further uses that to fine-tune the pre-trained model, but is still prone to performance degradation when the 2D keypoints are noisy. 
In contrast to prior arts, our proposed test-time adaptation method does not rely on 2D keypoints in any form.

\section{Method}

We first give a general overview of the proposed selective test-time adaptation ESCAPE framework, then define the human pose correction problem setup (\ref{sec:prob_form}) and describe the application of ESCAPE to this particular setting. A detailed overview of ESCAPE is shown in Fig. \ref{fig:framework}.

Our proposed method begins with the output prediction of a pre-trained backbone model given its input data. The energy score (\ref{sec:energy}) of this initial backbone prediction is compared against a predetermined threshold to classify the input data as out-of-distribution (OOD) or in-distribution (ID) for the backbone model. If the input data was ID, only a fast forward-pass correction (\ref{sec:cNet}) is applied. If instead the input data was OOD, the more intensive but powerful adaptive correction (\ref{sec:TTA}) is applied.

\begin{figure} [t]
    \centering
    \includegraphics[width=\textwidth]{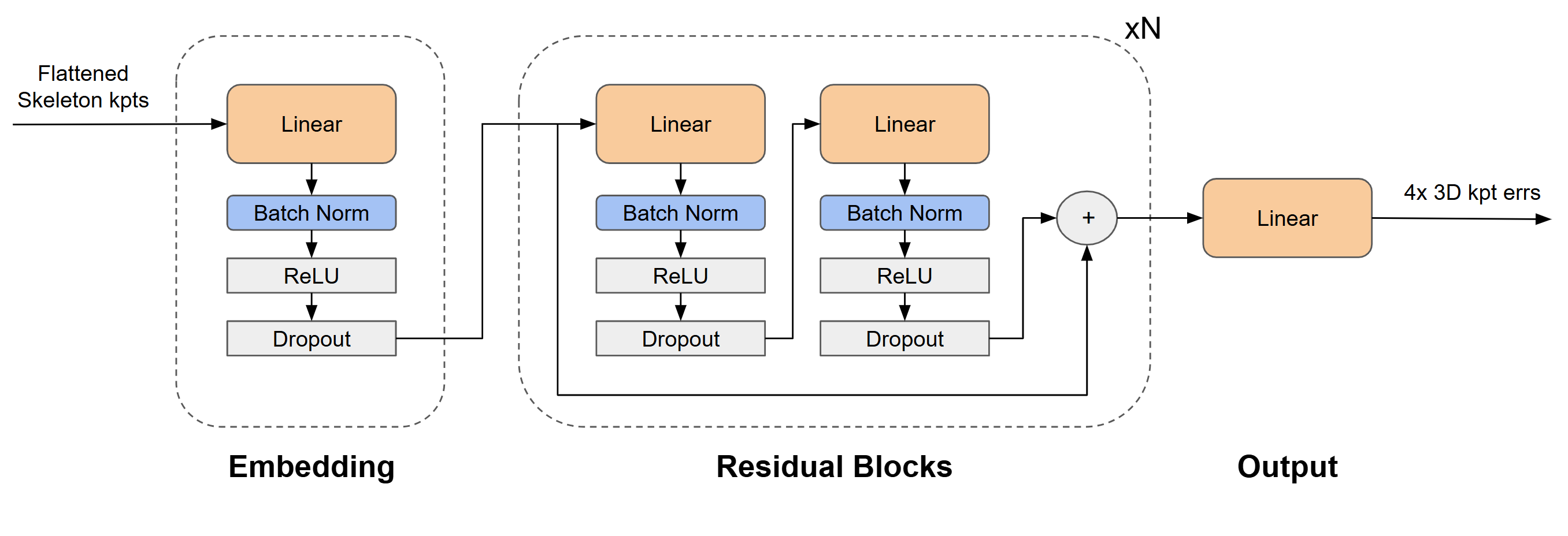}
    \caption{Diagram of the residual network architecture used for $\mathcal{C}$ and $\mathcal{R}$. It consists of an embedding module, a series of N residual blocks, and an output linear layer. The main building block is a linear layer followed by batch norm, ReLU activation, and dropout. The embedding consists of one such building block, while each residual block consists of two of these blocks in series, wrapped by a residual connection.}
    \label{fig:residual_arch}
\end{figure}

\subsection{Problem Formulation} \label{sec:prob_form}
    Given an input image frame $I$ of a person, a pre-trained human pose estimator $\mathcal{P}$ returns 3D human poses $X\in\mathbf{R}^{j\times3}$ where $j$ is the number of keypoints: 

    \begin{equation}
        \mathcal{P}(I) = X, \quad X\in\mathbf{R}^{j\times3}
        \label{eq:HPE}
    \end{equation}
    The proximal and distal keypoints $X_P$ and $X_D$ are subsets of the full keypoints $X$. The proximal keypoints $X_P$ are the keypoints at the base of each human limb: the right/left hips and the right/left shoulders. The distal keypoints $X_D$ are the keypoints at the end of each human limb: the right/left feet and the right/left hands. Estimating proximal joints is an easier task because of their greater bio-mechanical constraints, and human pose estimators have correspondingly lower estimation errors on proximal keypoints than on distal keypoints. We leverage this in the design of the reverse network test-time adaptation objective outlined in section \ref{sec:TTA}.

\subsection{Energy-based Sample Selection} \label{sec:energy}
To select hard samples from incoming data for inference time adaptation, we use the Energy Function, which has been shown to be effective in OOD detection in prior work \cite{energy,akbari2021ebjr,etran}.

Given the estimated human pose, $X\in\mathbf{R}^{j\times3}$, we flatten the estimated pose to create a vector, $f$, with the length of ${H=j\times3}$. 

We calculate the free energy over $f$ as follows:
\begin{equation}
    E(I)=-\log\sum_h^H e^{f^{(h)}(I)},
\end{equation}
where $I$ is the input image. 

We define a threshold, $E^{T}$,  to classify the input data $I$ as OOD or IND.
That is, if $E(I) < E^{T}$, we consider $I$ to be OOD for the backbone model, otherwise IND. 

\subsection{Correction Network} \label{sec:cNet}

    To improve backbone HPE predictions on the highly uncertain distal keypoints, we propose learning a simple correction network (CNet)  $\mathcal{C}$ to estimate the distal keypoint errors of input backbone pose predictions. The errors can then be directly applied to the distal keypoints of the backbone prediction to obtain a corrected estimation. That is,  $\mathcal{C}$  outputs estimated 3D errors for the $d$ distal keypoints $\Delta \hat{X}_{D} \in R^{d \times 3}$, given an input 3D human keypoint estimation $X$: 

    \begin{equation}
        \Delta \hat{X}_{D} = \mathcal{C}(X),
    \end{equation}
    and $X$ is subsequently adjusted using the predicted errors $\Delta \hat{X}_{D}$ to get a corrected pose $X^\mathcal{C}$. Note that $X^\mathcal{C}$ is identical to $X$ except for the adjusted distal keypoints:

    \begin{equation}
        X^\mathcal{C} = X - \Delta \hat{X}_{D}
    \end{equation}
    
    We train  $\mathcal{C}$  in a fully supervised manner, using the ground-truth 3D pose distal keypoints $Y_{D}$ from the training set of the backbone. 
    The distal error is defined as $\Delta X_{D} = Y_{D} - X_{D}$. The $L_2$ norm between estimated distal errors $\Delta \hat{X}_{D}$ and ground-truth distal errors $\Delta X_{D}$ is then used to train $\mathcal{C}$:
    
    \begin{equation}
        \mathcal{L}_{1} = ||\ \mathcal{C}(X),\Delta X_{D} ||_{2},
    \end{equation}

    While straightforward, using only $\mathcal{L}_1$ as supervision can overly encourage the network to address only scale issues in the initial backbone estimate. Since we equally want  $\mathcal{C}$  to correct orientation errors in the initial estimate, we define a second, closely related loss term $\mathcal{L}_{2}$ wherein $\mathcal{C}$ ingests the initial pose $X$ procrustes-aligned to the ground-truth pose $Y$, denoted by $X^{pa}$. The overall $\mathcal{L}_{2}$ term between the resulting prediction and the ground-truth error $\Delta X_{D}^{pa}=Y_{D} - X_{D}^{pa}$ is then defined as follows:

    \begin{equation}
        \mathcal{L}_{2} = ||\ \mathcal{C}(X^{pa}),\Delta X_{D}^{pa}||_{2}. 
    \end{equation}     

     Both $\mathcal{L}_{1}$ and $\mathcal{L}_{2}$ are combined into the final training objective for $\mathcal{C}$:

    \begin{equation}
        \mathcal{L}_\mathcal{C} = \lambda_1 \mathcal{L}_{1} + \lambda_2 \mathcal{L}_{2},
    \end{equation}
    where $\lambda_1$ and $\lambda_2$ are weights of the loss terms.

    At inference time, the trained $\mathcal{C}$ is used frozen on ID samples (in fast forward-pass correction) but is further tuned on OOD samples (in intensive adaptation).    

\subsection{Test-Time Adaptation via Self-Consistency} \label{sec:TTA}

    To adapt CNet to harder, OOD samples during inference, we define a Self-Consistency loss which uses only the backbone pose sample $X$, requiring no inference time ground-truth keypoints. This loss relies on a reverse network (RCNet), $\mathcal{R}$, which aims to learn the biomechanical reverse of the CNet task. $\mathcal{R}$  takes the corrected 3D human keypoint estimation $X^\mathcal{C}$ as input and outputs estimated 3D errors for each of the $p$ proximal keypoints $\Delta \hat{X}_{P} \in R^{p \times 3}$: 

    \begin{equation}
        \Delta \hat{X}_{P} = \mathcal{R}(X^\mathcal{C})
    \end{equation}

     At the training stage, we use ground truth 3D poses from proximal keypoints to train $\mathcal{R}$. 
    The proximal error is defined as $\Delta X_{P} = Y_{P} - X_{P}$. The $L_2$ norm between estimated proximal errors $\Delta \hat{X}_{P}$ and ground-truth distal errors $\Delta X_{P}$ is defined as:
    \begin{equation}
        \mathcal{L}_{\mathcal{R}} = ||\ \mathcal{R}(X),\Delta X_{D} ||_{2}.
    \end{equation}
    
    $\mathcal{R}$ is trained to predict proximal corrections given the pose corrected by $\mathcal{C}$ , $X^\mathcal{C}$. It only ever takes $X^\mathcal{C}$ as input.
    Learning $\mathcal{R}$ is easier than learning $\mathcal{C}$ since the proximal keypoints are more constrained compared with distal keypoints and therefore, once trained, $\mathcal{R}$ is kept frozen and used in TTA to adapt $\mathcal{C}$.
    
    To perform TTA on CNet using RCNet, the initial prediction $X$ is corrected by CNet. Then, the corrected sample $X^\mathcal{C}$ is input to  $\mathcal{R}$  resulting in a pose $X^{\mathcal{R}}$ with both corrected distal joints from  $\mathcal{C}$  and corrected proximal joints from  $\mathcal{R}$. The $L_{2}$ distance between the $X^{\mathcal{R}}$ proximal joints $X^{\mathcal{R}}_P$ and the original backbone pose proximal joints $X_P$ is used as the TTA objective $\mathcal{L}_{TT}$ to update  $\mathcal{C}$. Only $\mathcal{C}$ is updated using $\mathcal{L}_{TT}$ during adaptation, while $\mathcal{R}$ remains frozen. More formally, we define the test time consistency loss as:

    \begin{equation}
        \mathcal{L}_{TT} = ||\ X^{\mathcal{R}}_{P}, X_{P} ||_{2}.
    \end{equation}

\begin{figure} [t]
        \centering
        \includegraphics[scale=0.65]{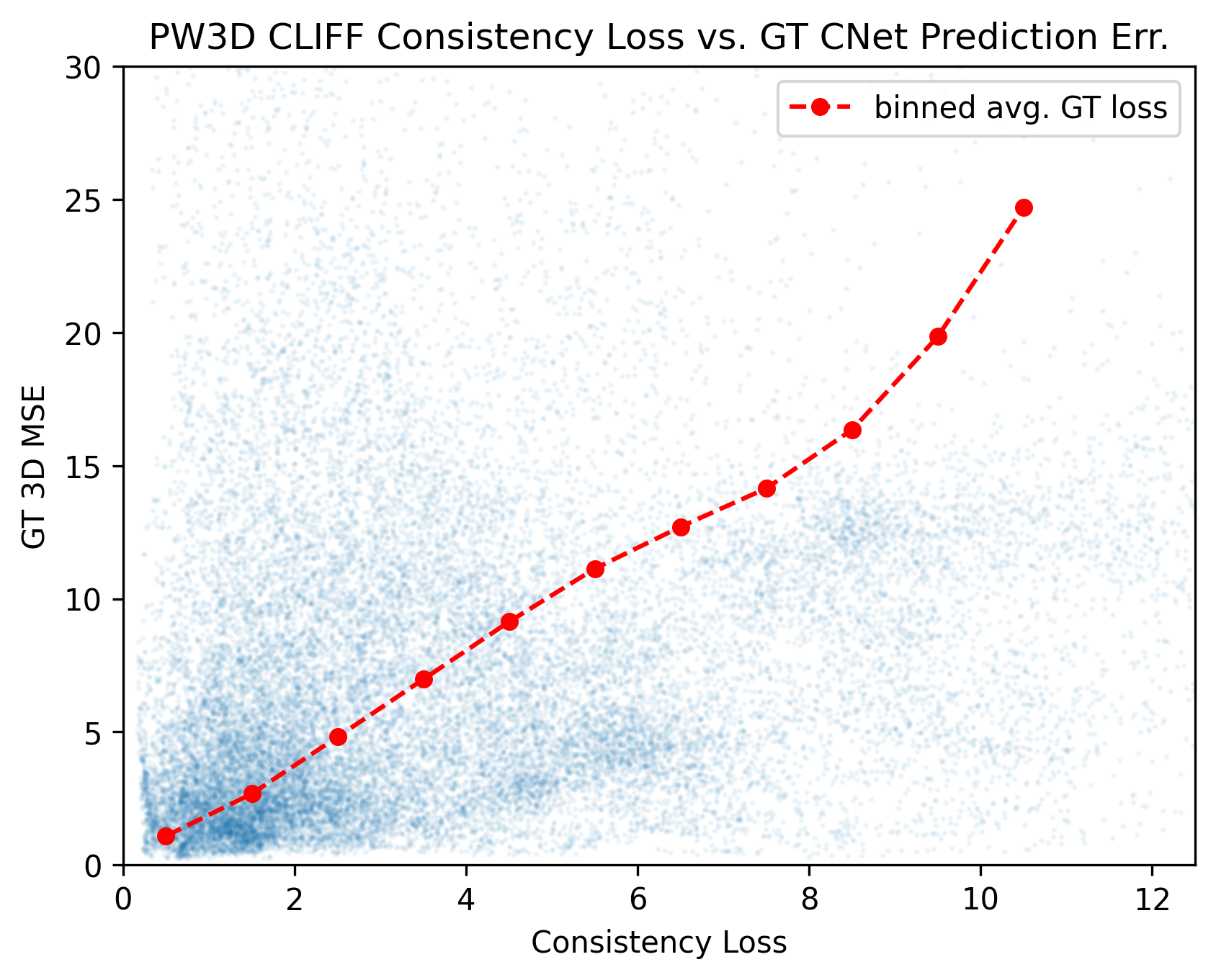}
        \caption{Strong correlation between the proposed self-consistency loss and the ground-truth 3D prediction error of CNet on 3DPW dataset, with CLIFF as the backbone pose estimator. Each blue point represents an individual test image, while binned averages are plotted in red.}
        \label{fig:loss_corr}
    \end{figure}

    In Fig. \ref{fig:loss_corr}, we empirically demonstrate that the proposed self-consistency loss is highly correlated with the ground-truth 3D mean squared error (MSE) by reporting the self-consistency loss and ground-truth MSE for every sample in the 3DPW test set \cite{vonMarcard2018} using CLIFF as the backbone network. Due to the large number of samples, we plot the binned average values in red. While individual samples may have a high degree of noise, the proposed TTA loss has a clear correlation with the ground-truth 3D objective.

    To further understand the proposed loss intuitively, first consider a sample $X^{\text{ind}}$ which lies ID for the backbone network. Because  $\mathcal{C}$  is trained using the backbone training data predictions, $X^{\text{ind}}$ will also be ID for $\mathcal{C}$. Similarly, it is ID for  $\mathcal{R}$  since $\mathcal{R}$ is trained using trained  $\mathcal{C}$  predictions on the backbone training data. If  $\mathcal{C}$  and  $\mathcal{R}$  have been trained well, $\mathcal{C}$'s correction to $X^{\text{ind}}$ and  $\mathcal{R}$ 's subsequent correction will both be minor since the sample was already good. Thus the distance between the distal joints of $X^{\text{ind}}$ and the twice-corrected pose will be small.
    
    In contrast, a sample OOD for the backbone network $X^{\text{ood}}$ will be OOD for  $\mathcal{C}$  and  $\mathcal{R}$ . Because it is OOD for the backbone, $X^{\text{ood}}$ will have unusually large error, receive more aggressive corrections, and result in greater distance between the proximals of the twice-corrected pose $X^{\mathcal{R}}$ and the uncorected pose $X$.
    
\section{Experiments}

In this section, we introduce the datasets, metrics and implementation details. We subsequently present a thorough evaluation of our proposed method, including experimental results, comparisons to state-of-the-art methods, and ablation studies on different parts of our framework.

\subsection{Experimental Setup} \label{sec:exp_setup}
    \textbf{Backbone Networks.} To demonstrate the general applicability of ESCAPE, we perform separate experiments using five widely-used and representative backbone HPE models: HybrIK \cite{hybrik}, SPIN \cite{spin}, PARE \cite{PARE}, BMSE \cite{bmse}, and CLIFF \cite{cliff}. Pose estimations from each backbone on the train and test data are produced and used to train and evaluate a separate pair of  $\mathcal{C}$  and  $\mathcal{R}$  for each backbone. 
    
    \textbf{Implementation Details.} The architecture of the residual networks used for $\mathcal{C}$ and $\mathcal{R}$ are identical, and follow that proposed by \cite{martinez2017simple}, which is often used as a 3D pose estimation baseline. A diagram of this architecture is shown in Fig. \ref{fig:residual_arch}. First, the flattened 3D skeleton keypoints are embedded via a linear layer followed by a batch norm layer, ReLU activation, and dropout layer. The embedding is passed through a series of residual blocks, then passed to a final linear layer to produce an output set of 4x 3D keypoint error vectors. We set all linear layer sizes = 512, use dropout=0.3, and use N=1 residual block. Our evaluation datasets use 17 keypoints in Human3.6M format, thus the network input shapes are both (B, 51).

    Our training procedure for $\mathcal{C}$ and $\mathcal{R}$ for a given pretrained backbone model is as follows: 1) we perform inference on the training images with the backbone model, recording the predicted and corresponding GT poses, 2) $\mathcal{C}$ is trained in a supervised manner using the predicted and GT poses with loss $\mathcal{L}_\mathcal{C}$ (eq. 7), 3) we perform inference on the training set of backbone predicted poses with the fully trained $\mathcal{C}$, recording the resulting distal-corrected poses $X^\mathcal{C}$, and finally 4) $\mathcal{R}$ is trained in a supervised manner using the distal-corrected poses $X^\mathcal{C}$ and GT poses with loss $\mathcal{L}_\mathcal{R}$ (eq. 9). Steps 1 and 2 can be easily combined, but we separated them in order to speed up the training of  $\mathcal{C}$ and $\mathcal{R}$ by avoiding repeated backbone model inference on the same images over the course of our work. Both are trained using the Adam optimizer for 30 epochs with batch size = 4096 using lr = 1e-4,  $\lambda_1$ = 1.0,  and $\lambda_2$ = 0.5. For TTA we take 2 steps with lr = 5e-4. 
    
    At inference, we perform per-sample adaptation and use test batch sizes = 1. We set the energy threshold = 800, and apply it to poses in millimetre scale. \textbf{Identical training and inference} hyperparameters are used across all backbones and datasets. No advanced hyperparameter searching or tuning was performed. An NVIDIA TITAN RTX GPU is used to produce all backbone pose estimations and to train and evaluate our networks.  

\subsection{Datasets}

To train  $\mathcal{C}$  and  $\mathcal{R}$ , we use the training splits of the MPII \cite{andriluka14cvpr} and 3DHP \cite{mono-3dhp2017} dataset. We limit ourselves to these datasets to demonstrate that ESCAPE requires no data other than that used to train the backbone in the first place. To evaluate ESCAPE, we follow prior works and use the 3DPW \cite{vonMarcard2018}, 3DHP, and SURREAL \cite{varol17_surreal} dataset test splits.

\begin{itemize}
    \item \textbf{3DHP} consists of in-the-wild and in-the-lab data from 8 subjects from 8 camera viewpoints. Six subjects (10.5k frames) make up the training set, two (2.9k frames) make up the test set. 
    \item \textbf{MPII} consists of a very wide range of everyday in-the-wild and indoor human activities extracted from YouTube videos (16.4k frames). While it is usually only used as a 2D dataset, we use pseudo-GT 3D annotations of MPII as the 3D dataset.
    \item \textbf{3DPW} is an in-the-wild dataset of subjects performing dynamic tasks including climbing, boxing, and playing basketball (35.5k frames). It contains many severe occlusions and has more dynamic camera poses than 3DHP.    
    \item \textbf{SURREAL} is a large-scale (6.5M frames) and realistic synthetic human pose estimation dataset containing 145 subjects with a large variety of poses, body shapes, clothings, viewpoints and backgrounds. The test set contains 30 of the subjects (1.2M frames).
\end{itemize}

\subsection{Evaluation Metrics}
Following previous works, we use the following metrics for evaluation: 1) mean per-joint position error (\textbf{MPJPE}) and 2) Procrustes-aligned mean per-joint position error (\textbf{PA-MPJPE}). Both metrics are measured in millimetres between the predicted and GT 3D coordinates, after root joint alignment.

\subsection{Quantitative Results}
    \textbf{3DPW Results.} In Tab. \ref{tab:3DPW}, we compare the all-keypoint performance of ESCAPE against state-of-the-art methods on the 3DPW test set. We report the distal performance improvement in Tab. \ref{tab:distals}. On 3DPW, ESCAPE improves the distal predictions of existing methods by significant margins: up to 6.3\% (Tab. \ref{tab:distals}), and improves the SOTA by up to 3.4\% on all-keypoints (Tab. \ref{tab:3DPW}).
    
    \textbf{3DHP Results.} Tab. \ref{tab:3DHP} presents results of ESCAPE and other state-of-the-art methods on the 3DHP test set. We again report the distal performance improvement in Tab. \ref{tab:distals}. ESCAPE improves distal predictions by up to 6.8\% on 3DHP (Tab. \ref{tab:distals}), and improves the SOTA by up to to 1.4\% over all-keypoints (Tab. \ref{tab:3DHP}).

    \textbf{SURREAL Results.} Tab. \ref{tab:SURREAL} presents results of ESCAPE and other state-of-the-art backbone methods on the SURREAL test set. We again report the distal performance improvement in Tab. \ref{tab:distals}. ESCAPE improves distal predictions by up to 4.0\% on SURREAL (Tab. \ref{tab:distals}), and improves the SOTA by up to to 1.4\% over all-keypoints (Tab. \ref{tab:SURREAL}).

\begin{table}
    \small
    \centering
    \caption{Results on 3DPW dataset. Models that are trained on the training set of 3DPW are shown with a checkmark.}
    \label{tab:3DPW}
    \begin{tabular}{ p{3cm}|c|ll}
        \toprule
        Method&FT&PA-MPJPE&MPJPE\\
        
        \midrule
        VIBE \cite{kocabas2020vibe} &\checkmark&51.9&82.9\\
        HybrIK \cite{hybrik} &\checkmark&41.8&71.3\\
        CLIFF \cite{cliff} &\checkmark&43.0&69.3\\
        \midrule
        BOA \cite{boa} &&78.7&138.8 \\
        DynBOA \cite{dynboa} &&77.0&142.0 \\
        DAPA \cite{weng2022domain} &&65.3&103.2\\
        SPIN \cite{spin} & &59.2&96.9\\
        PoseAug \cite{gong2021poseaug} &&58.5&94.1\\
        CycleAdapt \cite{CycleAdapt} &&53.9&85.8 \\
        HybrIK \cite{hybrik} &&49.3&88.7\\
        PARE \cite{PARE} &&49.4&81.8\\
        \midrule
        \textbf{Ours(+PARE)}&&\textbf{47.9}&\textbf{79.0}\\
        \bottomrule
    \end{tabular}
\end{table}

\begin{table}[t]
    \small
    \centering
    \caption{Results on 3DHP dataset. Models that are trained on the training set of 3DHP are shown with a checkmark.}
    \label{tab:3DHP}
    \begin{tabular}{ p{3cm}|c|ll}
        \toprule
        Method&FT&PA-MPJPE&MPJPE\\
        
        \midrule
        SPIN \cite{spin} & \checkmark&70.0&109.6\\
        AdaptPose \cite{adaptpose} &\checkmark&83.4&120.5\\
        PARE \cite{PARE} &\checkmark&69.3&102.3\\
        \midrule
        \textbf{Ours(+PARE)}&\checkmark&\textbf{69.2}&\textbf{100.9}\\
        \bottomrule
    \end{tabular}
\end{table}

\begin{table}[t]
    \small
    \centering
    \caption{Results on SURREAL dataset. None of the models are trained using any SURREAL data.}
    \label{tab:SURREAL}
    \begin{tabular}{ p{3cm}|c|ll}
        \toprule
        Method& & PA-MPJPE&MPJPE\\
        
        \midrule
        SPIN \cite{spin} & 94.3 & 131.9 \\
        BMSE \cite{bmse} & 96.8 & 149.8 \\
        PARE \cite{PARE} & 82.1 & 118.1 \\
        HybrIK \cite{hybrik} & 143.8 & 165.9 \\
        CLIFF \cite{cliff} & 86.1 & 126.7\\ 
        \midrule
        \textbf{Ours(+PARE)}&\textbf{82.1}&\textbf{116.4}\\
        
        \bottomrule
    \end{tabular}
\end{table}

\begin{table} [t]
    \centering
    \footnotesize
    \caption{Average single sample inference times on 3DPW for various test time adaptation methods using the SPIN (ResNet-50) backbone.}
    \label{tab:time_complexity}
    \begin{tabular}{l|c}
        \toprule
        \textbf{Method} & \textbf{Time (ms)} \\
        \hline
        
        SPIN \cite{spin} & 16.6 \\
        \hline
        \ + BOA \cite{boa} & 840.3 \\
        \ + DynBOA \cite{dynboa} & 1162.8 \\
        \ + DAPA \cite{weng2022domain} & 431.0 \\
        \ + CycleAdapt \cite{CycleAdapt} & 74.1 \\
        \hline
        \ + \textbf{Ours} & \textbf{23.0} \\
        \bottomrule
    \end{tabular}
\end{table}

\begin{table}
    \footnotesize
        \centering
        \caption{Distal keypoint improvements with various backbone estimators on the PW3D, HP3D, and SURREAL datasets}. ``Hard'' columns report average improvement over the backbone's top 10\% highest error samples, ``All'' columns report average on all samples.
        \label{tab:distals}
        \begin{tabular}{l|cc|cc|cc}
            \toprule
            & \multicolumn{2}{c}{\textbf{PW3D}} & \multicolumn{2}{c}{\textbf{HP3D}} & \multicolumn{2}{c}{\textbf{SURREAL}} \\
            & \multicolumn{2}{c}{\textbf{Distal MPJPE}} & \multicolumn{2}{c}{\textbf{Distal MPJPE}} & \multicolumn{2}{c}{\textbf{Distal MPJPE}} \\
            & All & Hard & All & Hard & All & Hard \\
            \hline
            {HybrIK}  & 134.1 & 204.8 & 143.5 & 258.8 & 161.6 & 375.9\\
            \ \ \ + \textbf{Ours} & \textbf{-5.8} & \textbf{-12.1} & \textbf{-9.8} & \textbf{-8.5} & \textbf{-1.5} & \textbf{-0.8}  \\
            \hline

            {BMSE}  & 138.9 & 255.7 & 157.6 & 319.8 & 186.4 & 398.3 \\
            \ \ \ + \textbf{Ours}& \textbf{-0.9} & \textbf{-7.25} & \textbf{-5.5} & \textbf{-6.3} & \textbf{-5.9} & \textbf{-10.2} \\
            \hline
            
            {SPIN}  & 140.7 & 238.6 & 153.4 & 312.8 & 171.9 & 350.8  \\
            \ \ \ + \textbf{Ours} & \textbf{-2.1} & \textbf{-8.5} & \textbf{-6.2} & \textbf{-6.3} & \textbf{-4.0} & \textbf{-4.0}\\
            \hline
            
            {PARE}  & 128.5 & 203.3 & 144.5 & 272.0 & 150.6 & 311.9 \\
            \ \ \ + \textbf{Ours}& \textbf{-8.1} & \textbf{-4.3} & \textbf{-4.9} & \textbf{-3.4} & \textbf{-6.0} & \textbf{-6.5} \\
            \hline

            {CLIFF} & 118.4 & 207.9 & 139.4 & 296.3 & 161.6 & 355.1\\
            \ \ \ + \textbf{Ours}& \textbf{-4.3} & \textbf{-5.7} & \textbf{-3.4} & \textbf{-3.9} & \textbf{-2.9} & \textbf{-7.6} \\ 
            \bottomrule
        \end{tabular}
\end{table}

\textbf{Inference Time.} In Tab. \ref{tab:time_complexity} we show that ESCAPE takes the least computation time compared to existing test-time adaptation methods. \textbf{We observe a 50x speedup over DynaBOA, 37x over BOA, 19x over DAPA, and 3x over the recent CycleAdapt,} which is the fastest of the prior arts by an order of magnitude. Additionally, in practice where no ground-truth 2D poses are available, the inference times of existing methods will be increased by needing to compute estimated 2D keypoints for supervision. ESCAPE uses no 2D pose supervision at all and so does not have this drawback. BOA \cite{boa} and DynaBOA \cite{dynboa} have massive inference times due to performing two network updates on every image. The inference time of DAPA \cite{weng2022domain} suffers from rendering an image of a synthetic 3D pose for every image. While CycleAdapt \cite{CycleAdapt} and the other methods tune all parameters in the large pretrained backbone during adaptation to every encountered sample, ESCAPE only adapts to hard samples and only tunes the small correction network $\mathcal{C}$ during adaptation, leaving the backbone frozen.

\subsection{Qualitative Results} \label{sec:qualitative_results}
\begin{figure}[t]
    \centering
    \includegraphics[scale=0.5]{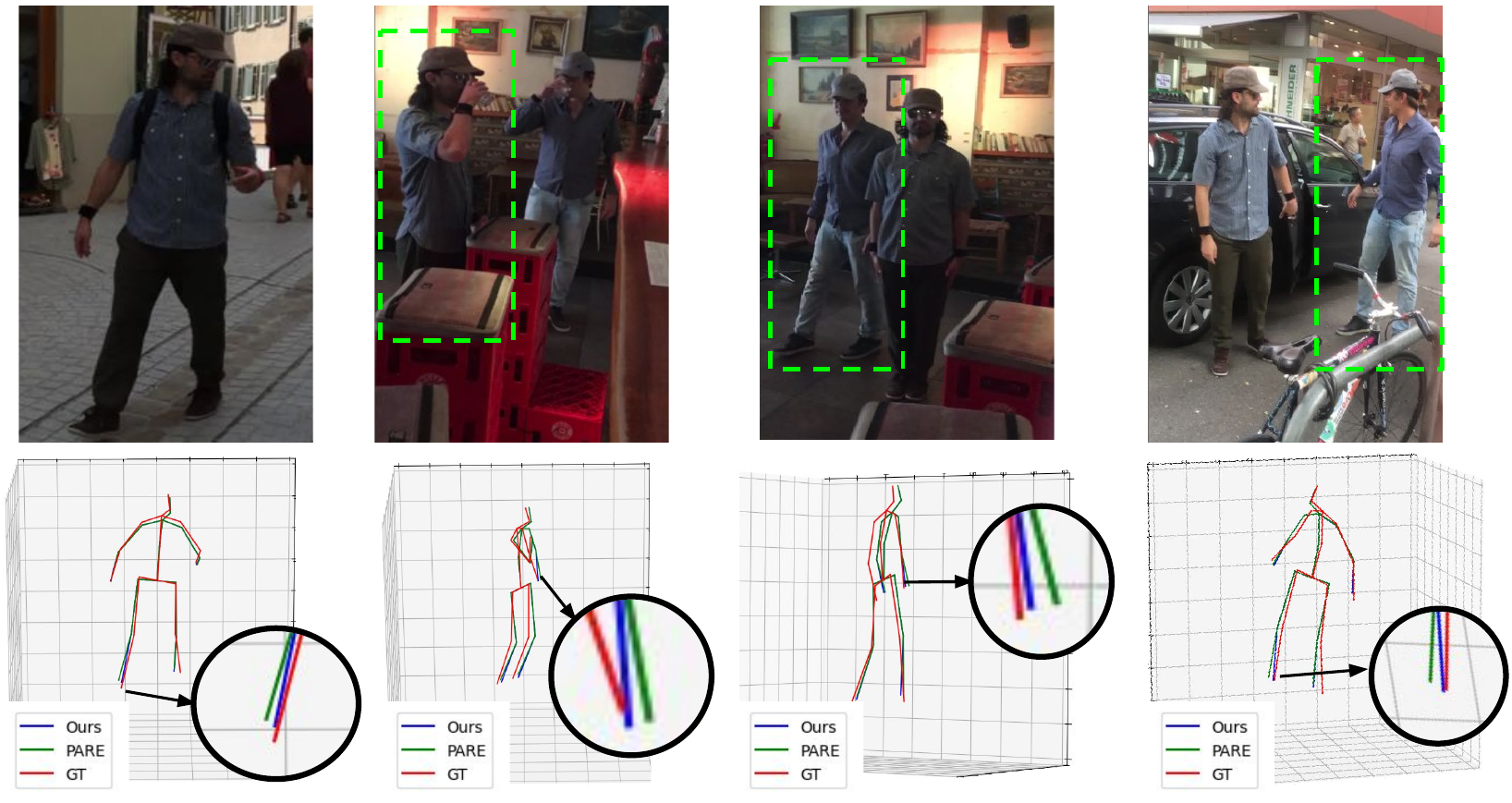}
    \caption{Example ESCAPE corrections to backbone predictions on samples from 3DPW. The top row shows images input to the backbone estimator and the bottom row shows the GT, backbone predicted, and corrected backbone predicted 3D poses.}
    \label{fig:qualitative}
\end{figure}

Fig. \ref{fig:qualitative} presents a qualitative evaluation of corrections made to PARE backbone predictions on samples from 3DPW. Input images and the corresponding GT, initial backbone prediction, and distal-corrected predictions from ESCAPE are shown for each. Even though the backbone predictions are already quite good, ESCAPE is able to significantly correct their distal keypoints to be closer to those of the ground-truth pose. 

\begin{figure} [t]
    \centering
    \includegraphics[scale=0.25]{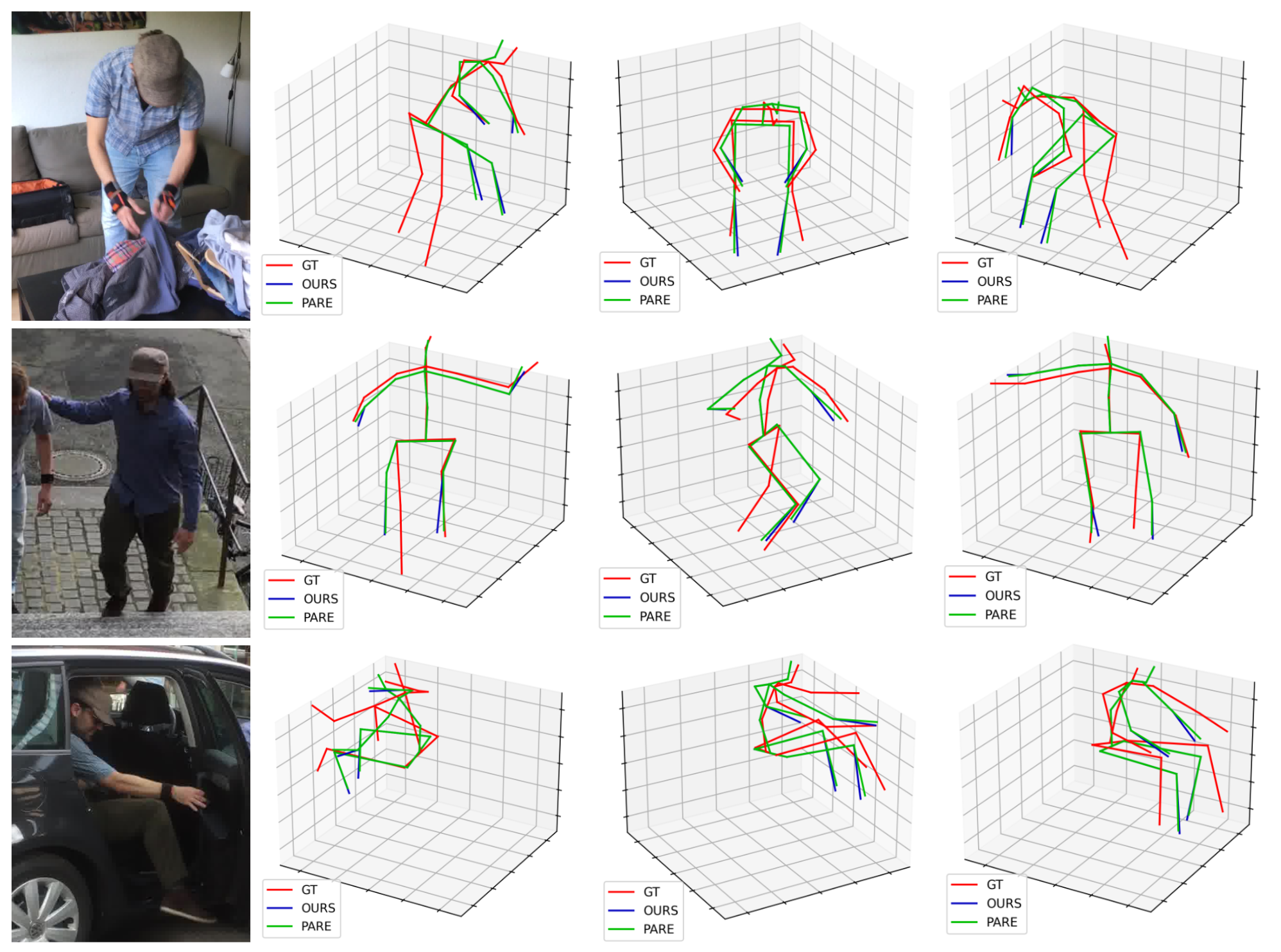}
    \caption{Example failure cases of ESCAPE. The leftmost column shows the input image, while columns 2,3 and 4 show three views of the GT (red), backbone estimator (green), and ESCAPE (blue) 3D poses.}
    \label{fig:failure_cases}
\end{figure}

\textbf{Failure Cases.} While the proposed method shows promising results, there are instances in which it is unable to succesfully improve the backbone prediction. To investigate the failure modes of ESCAPE, we consider its worst performing samples (i.e., with the greatest increase in MPJPE/PA-MPJPE of corrected pose vs backbone estimation) and specifically noted two major categories of backbone estimation mistakes which ESCAPE has difficulty correcting: 1) major overall pose (torso) misalignment, and 2) major simultaneous distal and near-distal keypoint mistakes (elbows, knees), both of which are due to upstream mistakes made by the backbone model. In Fig. \ref{fig:failure_cases} we illustrate representative failure cases for ESCAPE, selected from the worst performing samples on PW3D using the PARE backbone. For each example, the input image (1st column) and corresponding GT, initial backbone prediction, and distal-corrected predictions from ESCAPE (the 2nd, 3rd, 4th columns) are shown.

 The first category primarily affects the corrected pose MPJPE but may still result in improved PA-MPJPE. In this case, the estimated pose has a significant misalignment w.r.t the GT pose (most often via torso orientation error) so even if the estimated pose has generally correct inter-keypoint relationships, a well-trained ESCAPE will make corrections which are reasonable within the mistaken orientation but which increase the error relative to the GT orientation. The second category is more challenging since it generally causes an irrecoverable loss in information for the affected limb. Considering the second example in Fig. \ref{fig:failure_cases}, if both the left knee and left ankle keypoints have major mistakes which changes a stair-climbing stride to a firmly planted crouch, ESCAPE can not successfully predict the real GT pose since it only has the estimated pose without any other context or input.

While these mistakes highlight some of the drawbacks stemming from the low-dimensional keypoint input of ESCAPE, the advantages in simplicity and speed make it a worthwhile trade-off. Additionally, it should be noted that the proposed corrections maintain reasonable poses even in these most difficult scenarios.

\subsection{Ablation Studies} \label{sec:ablations}

\textbf{Effectiveness of energy threshold in selecting hard samples.} We measure the average backbone prediction MPJPE over each full dataset (Tab. \ref{tab:E_thresh_top10}, column All) and over each corresponding subset of energy-selected samples (Tab. \ref{tab:E_thresh_top10}, column OOD). 
Additionally, we measure the fraction of the top 10\% error samples remaining after energy thresholding. 
The results are summarized in Tab. \ref{tab:E_thresh_top10}. 
The same energy threshold is used across all datasets and backbones. 
For all datasets and backbones, we observe that the average MPJPE of the energy threshold selected samples is always higher than that of the full set of samples, indicating that the energy threshold effectively keeps hard (OOD) samples while rejecting easier (ID) samples. 
In the last column of Tab. \ref{tab:E_thresh_top10} we observe that despite using the same E threshold value across all backbones and datasets, more than 60\% of the top 10\% samples pass the energy threshold on the 3DHP dataset. This is comparable with prior work using the energy function \cite{energy} which found that even in classification, roughly 40\% of OOD and ID samples were overlapped.

\begin{table} [t]
        \centering
        \small
        \caption{Ablation study investigating the energy function hard sample separation across multiple backbones and datasets. For each backbone model and dataset, the average MPJPE of all samples and of those passing the energy threshold are reported in the ``All'' and ``OOD'' columns respectively. The fraction of top 10\% error samples remaining is reported in the ``Acc'' column. Datasets used to train the backbones are marked with a check.}
        \label{tab:E_thresh_top10}
        \begin{tabular}{l|l|c|cc|c}
            \toprule
            \textbf{Model} & \textbf{Dataset} & \textbf{Train} & \textbf{All} & \textbf{OOD} & \textbf{Acc}\\
            \hline
            \textbf{HybrIK} & 3DPW && 87.9 & 88.2 & 0.16\\
                  & 3DHP(test) && 107.4 & 107.9 & 0.53 \\
                  & MPII & \checkmark & 113.3 & 129.9 & 0.78 \\
                  & 3DHP(train) & \checkmark & 87.3 & 89.5 & 0.73 \\
            \hline
            \textbf{SPIN} & 3DPW && 94.1 & 112.8 & 0.45\\
                  & 3DHP(test) && 109.6 & 124.8 & 0.67 \\
                  & MPII & \checkmark & 138.3 & 160.7 & 0.91 \\
                  & 3DHP(train) & \checkmark & 70.4 & 74.7 & 0.69 \\
            \hline
            \textbf{PARE} & 3DPW && 81.8 & 87.0 & 0.58\\
                  & 3DHP(test) && 102.3 & 105.4 & 0.70 \\
                  & MPII & \checkmark & 114.1 & 122.3 & 0.92 \\
                  & 3DHP(train) & \checkmark & 115.5 & 122.5 & 0.88 \\
            \hline
            \textbf{BMSE} & 3DPW && 93,9 & 122.0 &0.61\\
                  & 3DHP(test) && 110.5 & 127.1 &0.56 \\
                  & MPII & \checkmark & 124.6 & 144.3 & 0.88 \\
                  & 3DHP(train) & \checkmark & 90.4 & 98.0 & 0.65 \\
            \hline
            \textbf{CLIFF} & 3DPW && 76.5 & 89.5 &0.67\\
                  & 3DHP(test) && 99.6 & 106.1 & 0.75 \\
                  & MPII & \checkmark & 89.3 & 93.7 & 0.95 \\
                  & 3DHP(train) & \checkmark & 91.2 & 96.0 & 0.86 \\
            \bottomrule
        \end{tabular}
    \end{table}

\textbf{Components of ESCAPE.} In Tab. \ref{tab:ab_select} we show the effects of the main components of ESCAPE including CNet, TTA, and Energy-based sample selection ($E_{T}$) on the performance of five pre-trained HPE models. Applying intensive adaptation to all samples is more powerful than only applying the fast correction, but only adapting to hard samples selected by the energy threshold and otherwise applying fast correction maintains or further improves performance while simultaneously cutting the average inference time overhead by 60\%. The unmodified backbone performance is shown in the first row of each block, followed by performance improvements from 1) applying the fast correction network $\mathcal{C}$ to all backbone samples, 2) applying the intensive self-consistency TTA to all backbone samples, and 3) applying full ESCAPE: selectively applying intensive self-consistency TTA to hard samples while only applying fast correction otherwise. In the third column (Fast) we report the number of samples to which only fast correction is applied. These findings demonstrate the effectiveness of each component of the proposed framework. 

We also show the number of test samples which are adapted to in the fourth column (TTA). The last column reports the average increase in inference time introduced on top of the inference time of the backbone model. Selecting OOD samples for test time adaptation significantly decreases the number of test samples that were used for adaptation (more than 50\%) and lowers the inference time by about 60\% while the improvements remain unchanged.

    \begin{table} [H]
    \small
        \centering
        \small
        \caption{Multi-backbone ablation study of ESCAPE components on 3DPW. We measure the improvement of fast correction on all samples (+CNet), intensive adaptation on all samples (+TTA), and selective adaptation (+E\_T) over the baseline method. Improvement on all samples and on backbone's top 10\% highest error samples is reported in ``all'' and ``Hard'' columns respectively. The final column reports average inference time increase.
        }
        \begin{tabular}{l|cc|cc|c}
            \toprule
            &\multicolumn{2}{c}{\textbf{MPJPE}} & \multicolumn{2}{c}{\textbf{num samples}} & \\
             \textbf{Method} & \textbf{All} & Hard & Fast & TTA & $\Delta$T (ms)\\
            \hline
            \textbf{HybrIK}  & 87.9 & 147.3 && \\
            \ \ \ +CNet & -1.6 & -3.2 & 35k & 0 & 1.0 \\
            \ \ \ +TTA & \textbf{-1.7} & \textbf{-3.6} & 0 & 35k & 27.6 \\
            \ \ \ +E\_T & -1.6 & -3.4 & 28.4k & 5.6k & 5.2 \\

            \hline
            \textbf{SPIN}    & 94.1 & 175.3 && \\
            \ \ \ +CNet & -0.6 & \textbf{-2.7} & 35k & 0 & 1.0 \\
            \ \ \ +TTA  & \textbf{-0.7} & -2.4 & 0 & 35k & 27.6 \\
            \ \ \ +E\_T & -0.6 & -2.1 & 27.9k & 7.1k & 6.4 \\

            \hline
            \textbf{PARE}  & 81.8 & 140.9 && \\
            \ \ \ +CNet & -2.2 & -1.6 & 35k & 0 & 1.0 \\
            \ \ \ +TTA & \textbf{-2.8} & \textbf{-1.9} & 0 & 35k & 27.6 \\
            \ \ \ +E\_T & -2.6 & \textbf{-1.9} & 18k & 17k & 13.9 \\
            
            \hline
            \textbf{BMSE} & 93.9 & 186.5 && \\
            \ \ \ +CNet & 0.0 & -1.5 & 35k & 0 & 1.0 \\
            \ \ \ +TTA & \textbf{-0.3} & -1.7 & 0 & 35k & 27.6 \\
            \ \ \ +E\_T & \textbf{-0.3} & \textbf{-1.9} & 26.6k & 8.4k & 7.4 \\
            
            \hline
            \textbf{CLIFF}    & 76.5 & 147.9 && \\
            \ \ \ +CNet & -1.1 & -0.8 & 35k & 0 & 1.0 \\
            \ \ \ +TTA  & \textbf{-1.2} & -1.0 & 0 & 35k & 27.6 \\
            \ \ \ +E\_T & \textbf{-1.2} & \textbf{-1.1} & 21.5k & 13.5k & 11.3 \\
            \bottomrule
            
        \end{tabular}
        \label{tab:ab_select}
    \end{table}
    
\textbf{Alternative OOD selection method.} While we have proposed using the Energy function for OOD sample selection in our framework, other methods can easily be used in its place. Since the accuracy of the OOD sample selection method has a great impact on the effectiveness of ESCAPE, we investigate the use of a recent OOD detection method, NNGuide \cite{10377411}, in place of the Energy function; and compare the results for PW3D and HP3D are in Table \ref{tab:NNGuide}. We report the MPJPE of the distal keypoints in the ``Distal MPJPE'' columns and the number of samples selected for intensive adaptation in the ``\# TTA'' columns. An NNGuide OOD score threshold of 63 was empirically determined such that the number of selected samples is comparable to the Energy function, and we use $k = 100$ since our training data is not too large. We note that the results are robust to the two studied OOD detection methods (as indicated by overall comparable performances), despite that NNGuide requires some learning from training samples.

\begin{table}[]
        \centering
        \begin{tabular}{l|cc|cc}
        \toprule
            & \multicolumn{2}{c}{PW3D} & \multicolumn{2}{c}{HP3D} \\

              & Distal MPJPE & \# TTA & Distal MPJPE & \# TTA \\ 
            \hline
            PARE & 122.2 & - & 144.5 & - \\
            \ \ \ +Ours (Energy) & -8.1 & 17k & \textbf{-4.9} & 2k \\ 
            \ \ \ +Ours (NNGuide) & \textbf{-9.3} & 12.6k & \textbf{-4.9} & 1.7k \\ 
            \hline
            HybrIK & 134.1 & - & 143.5 & - \\
            \ \ \ +Ours (Energy) & \textbf{-5.8} & 5.6k & -9.8 & 1.4k \\
            \ \ \ +Ours (NNGuide) & \textbf{-5.8} & 9k & \textbf{-10.2} & 1.3k \\
            \hline
            BMSE & 138.9 & - & 157.6 & - \\
            \ \ \ +Ours (Energy) & \textbf{-0.9} & 8.4k & \textbf{-5.5} & 1.2k  \\
            \ \ \ +Ours (NNGuide) & 1.7 & 8.1k & -5.3 & 1.2k \\
            \hline
            SPIN & 140.7 & - &  153.4 & - \\
            \ \ \ +Ours (Energy) & \textbf{-2.1} & 7.1k & \textbf{-6.2} & 1.3k \\
            \ \ \ +Ours (NNGuide) & -1.9 & 7.4k & -5.9 & 1.3k \\
            \hline
            CLIFF & 118.4 & - & 139.4 & - \\
            \ \ \ +Ours (Energy) & -\textbf{4.3} & 13.5k & -3.4 & 1.9k\\
            \ \ \ +Ours (NNGuide) & -3.9 & 13.9k & \textbf{-5.8} & 1.6k \\
            
        \bottomrule
        \end{tabular}
        \caption{Distal MPJPE improvements (Distal MPJPE) and the number of samples selected for TTA (\# TTA) of ESCAPE on PW3D and HP3D, when using NNGuide with the threshold of 63 (i.e., +Ours (NNGuide)) as the OOD detection method vs. Energy function with threshold of 800 (i.e., +Ours (Energy)).}
        \label{tab:NNGuide}
    \end{table}

\textbf{Energy based sample selection with BOA.} The use of the energy function to select hard samples for adaptation is not limited to our framework but can be used to make existing TTA methods more efficient and practical. By choosing hard, OOD samples to adapt to instead of adapting blindly to all incoming samples, the large computational overhead of TTA methods can be mitigated while retaining most of their improvement. To evaluate this claim, we investigate the effectiveness of using our energy method to select samples for adaptation with BOA.
    
In Tab. \ref{tab:sup_BOA} we report the results of using BOA to adapt a pretrained HMR \cite{kanazawa2018end} network to 3DPW on all samples (+BOA), and by selecting samples for adaptation using our proposed Energy-based method with different Energy threshold values (+E800, +E750, +E700). We also compare against a naive random baseline which selects incoming samples with probability equal to the fraction of samples selected by the Energy threshold. That is, such that its number of selected samples is comparable to that selected by the Energy threshold. We report results with Energy threshold values of 800, 750, and 700 and include corresponding random baselines for each threshold value (+Rand). 

\begin{table} [t]
\small
    \centering
    \small
    \caption{Comparison of different sample selection strategies for test-time adaptation of HMR with BOA on 3DPW. Results with multiple Energy threshold values are reported, with corresponding random baselines. The last two columns report the number of samples selected for adaptation and the factor by which average inference time is increased compared to HMR without adaptation.
    }
    \begin{tabular}{l|ccc|c|c}
        \toprule
        \textbf{Method} & PA-MPJPE & MPJPE & PVE & \# adapt & Time \\
        
        \midrule
        \textbf{HMR} & 66.4 & 109.6 & 131.0 & - & - \\
        \ \ \ +BOA  & \textbf{55.7} & \textbf{89.0} & \textbf{108.0} & 35k & 51 x \\   

        \midrule
        \ \ \ +E800 & 61.1 & 96.7 & 116.4 & 7.5k & 12 x \\ 
        \ \ \ +Rand & 60.8 & 99.3 & 119.4 & 7.5k & 12 x \\    

        \midrule
        \ \ \ +E750 & 62.4 & 98.8 & 118.9 & 4.1k & 7 x \\ 
        \ \ \ +Rand & 62.8 & 102.4 & 123.0 & 4.1k & 7 x \\    

        \midrule
        \ \ \ +E700 & 63.2 & 100.1 & 120.5 & 3.3k & \textbf{6 x} \\ 
        \ \ \ +Rand & 63.2 & 103.2 & 124.0 & 3.3k & \textbf{6 x} \\    
        
        \hline
    \end{tabular}
    \label{tab:sup_BOA}
\end{table}

There are a few differences in our setup compared to the original BOA setup which should be noted. It is common for existing adaptation works, including BOA, to demonstrate the capabilities of their method in adapting a poorly trained model to 3DPW. Generally, this means pre-training the backbone pose estimator only on the H36M \cite{ionescu2013human3} dataset, resulting in an impractical network that suffers from unrealistically large domain shifts at inference. While this is a useful approach for demonstration, we aim to show the benefit of our Energy selection method in a realistic setting, where the backbone has been trained on a variety of common datasets and shows good generally performance, but would still benefit from closing the train-test distribution gap. Instead of just H36M, we pretrain the same HMR backbone on a more realistic collection of datasets: H36M, 3DHP \cite{mono-3dhp2017}, LSP \cite{johnson2010clustered}, lspet \cite{johnson2011learning}, MPII \cite{andriluka14cvpr}, and COCO 2014 \cite{lin2014microsoft}. We use the implementation provided by \cite{mmhuman3d} for this. 

We do not use BOA's temporal-wise losses since by selecting poses for adaptation according to the Energy threshold or random sampling, we no longer have a fixed interval between frames. Additionally, we found the original hyperparameters of BOA to be unsuitable for the realistically trained network and reduced the number of adaptation steps from 7 to 3, reduced the lower-level fast learning rate from 8e-6 to 4e-6, and reduced the upper-level learning rate from 3e-6 to 1e-6. 

We find that adapting to only a fraction of incoming samples can keep most of the improvement gained by adapting to all samples. \textbf{Using an Energy threshold of 750 to select samples for adaptation with BOA has a 7x faster average inference time than adapting to all samples, while keeping the majority of the improvement.} Naively adapting to all incoming samples with BOA increases the inference time by 51x, while only adapting to selected hard samples results in far more modest 6x, 7x, and 12x increases for energy thresholds of 700, 750, and 800. As in ESCAPE, applying intensive adaptation to all samples encountered provides the best performance but suffers the largest computational overhead. Importantly, the Energy-based selection strategy always performs better than the random baselines. We note that the results further suggest that adapting to all incoming samples, as BOA and other methods do, is highly inefficient, since randomly sampling incoming samples without any strategy is also surprisingly effective. Overall, these results support the validity of our assumptions and demonstrate the ability of the proposed Energy-based sample selection to improve the efficiency of existing test-time adaptation methods.

\textbf{Impact of Varying Conditions.} In many real-world scenes, challenges such as occlusions and varying lighting can significantly degrade the performance of pose estimation models. To systemically investigate the impact of these variations on the proposed method, we compile a new evaluation dataset comprising of 1000 images randomly selected from all PW3D testset samples on which ESCAPE performed reasonably well (where the CNet correction gives improvement in PA-MPJPE/MPJPE over the backbone prediction, and CNet after TTA also gives improvement in PA-MPJPE/MPJPE over the un-adapted CNet correction). We then introduce synthetic variations in lighting and occlusion to create seperate variations of the new dataset and evaluate the resulting performance of ESCAPE in order to understand their impact on the performance. 

\begin{figure} [H]
    \centering
    \includegraphics[width=0.75\textwidth]{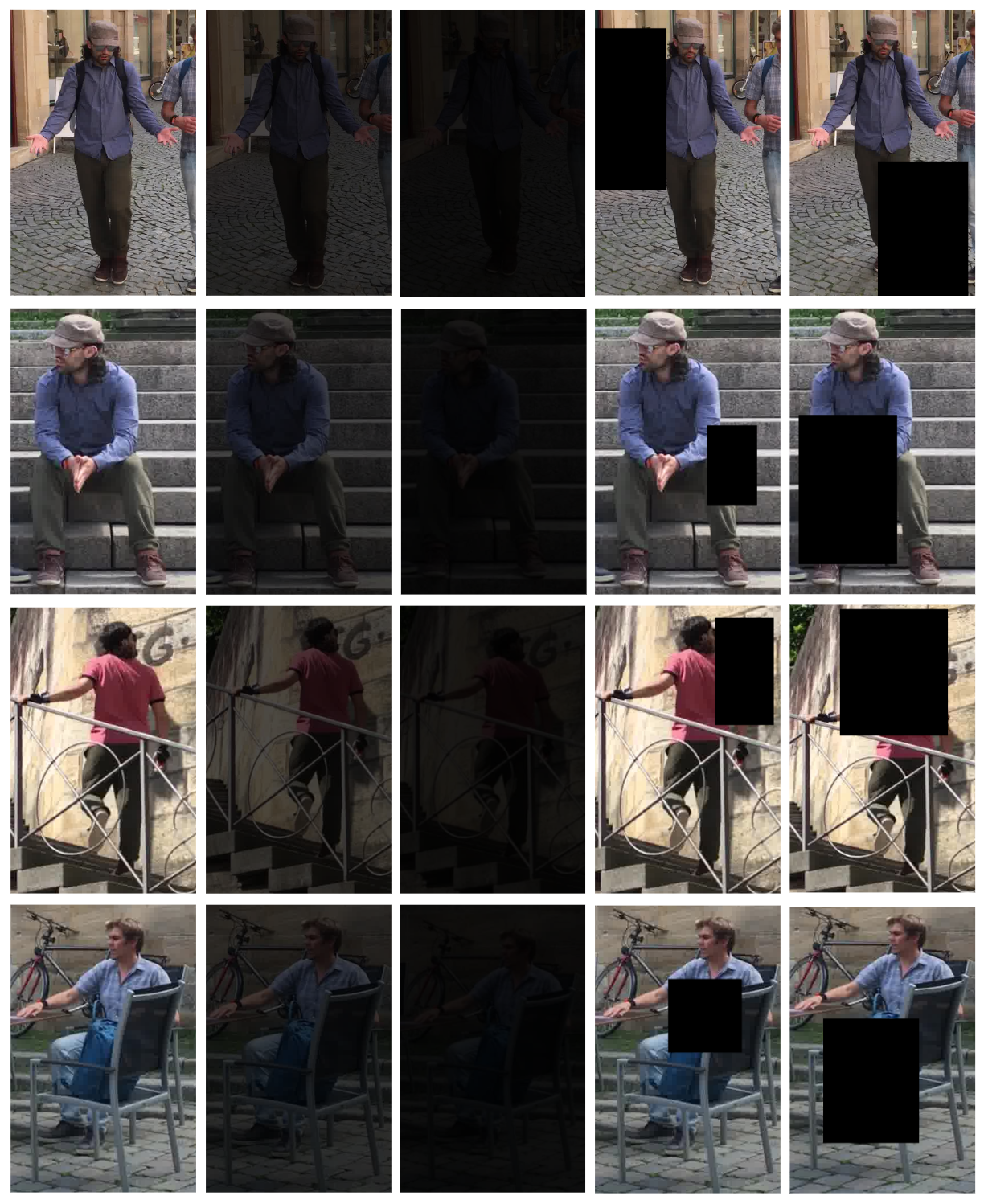}
    \caption{Example images of artificial varying conditions from PW3D. The 1st column shows the original image and the folowing columns from L to R show darkening (alpha=0.3), darkening (alpha=0.1), occlusion mild, and occlusion moderate variations.}
    \label{fig:varying_conditions_samples}
\end{figure}

To simulate lighting variations, we simply blend the original image with a fully black image, with ratios of original/black of 0.3 and 0.1. To simulate occlusions, we overlay randomly sized black rectangles at random locations in the ROI defined by the GT pose bounding box. We define a `Mild' setting in which the random rectangle has width ranging from 0.4x and 0.7x the bounding box width and height ranging from 0.3x to 0.6x the bounding box height. We also define a `Moderate' setting placing a rectangle with width ranging from 0.6x and 0.8x the bounding box width and with height ranging from 0.5x to 0.8x the bounding box height. Fig. \ref{fig:varying_conditions_samples} provides some examples of the raw images and corresponding variations.

Fig. \ref{fig:varying_conditions_samples} presents some examples of the raw and synthesized samples and we summarize the resulting performance of ESCAPE with PARE backbone on the raw images and the distorted images under 4 variation settings in Tab. \ref{tab:varying_conditions}. We report both the performance of the fast CNet correction on all samples (+CNet only) and also the performance of the full proposed method with TTA applied to samples selected by the energy function threshold (+ESCAPE). The table shows that, as expected, the performance of ESCAPE is good on the raw images and that the performance of the backbone (PARE) is degraded by both increasingly darkened and increasingly occluded images. On this dataset, we report the results averaged over N=100 trials to counteract stochasticity in the TTA update on a single sample.

Despite the challenging settings, we find that CNet remains able to significantly improve the backbone predictions and the self-consistency TTA remains able to improve CNet to a similar degree, demonstrating the robustness of ESCAPE. However, we note that the improvements degrade along with the backbone predictions and that occlussions in particular become more challenging for both the backbone and ESCAPE, as they increase from Mild to Moderate. As discussed in section \ref{sec:qualitative_results}, we identify two different major categories of backbone mistakes which ESCAPE can have difficulty correcting, both of which often result from challenging distal occlusions: 1) major overall pose (torso) misalignment, and 2) major simultaneous distal and near-distal keypoint mistakes (elbows, knees). We present some examples of these backbone mistakes and the corresponding ESCAPE corrections in Fig. \ref{fig:failure_cases}).

Overall, we observe that despite the large variations in actions, camera position, and lighting, ESCAPE is able to noticeably improve the backbone predictions.

\begin{table} [t]
    \small
        \centering
        \small
        \caption{Results under varying conditions on the 1000-sample subset of PW3D when using the PARE backbone. The results are averaged over N=1000 trials.}
        \begin{tabular}{c|l|cc}
                \toprule
                \textbf{Variation} & \textbf{Method} & \textbf{PA-MPJPE} & \textbf{MPJPE}\\
                \hline
                \ \ \textbf{Raw} & PARE & 51.9 & 86.0\\
                \ \ \ & +CNet only & \textit{-2.8} & \textit{-3.4} \\
                \ \ \ & +ESCAPE & \textit{-3.1} & \textit{-4.4} \\
                \hline
                \ \ \textbf{Dark 0.3} & PARE & 52.0 & 87.8\\
                \ \ \ & +CNet only & \textit{-2.1} & \textit{-3.0} \\
                \ \ \ & +ESCAPE & \textit{-2.0} & \textit{-3.7} \\
                \hline
                \ \ \textbf{Dark 0.1} & PARE & 56.8 & 100.5 \\
                \ \ \ & +CNet only & \textit{-1.2} & \textit{-2.0} \\
                \ \ \ & +ESCAPE & \textit{-0.8} & \textit{-2.4} \\
                
                \hline
                \ \ \textbf{Occl. Mild} & PARE & 58.9 & 96.1 \\
                \ \ \ &+CNet only& \textit{-2.4} & \textit{-2.8} \\
                \ \ \ &+ESCAPE & \textit{-2.8} & \textit{-3.7} \\
                \hline
                \ \ \textbf{Occl. Moderate} & PARE & 65.5 & 106.5 \\
                \ \ \ &+CNet only& \textit{-2.1} & \textit{-2.5} \\
                \ \ \ &+ESCAPE & \textit{-2.4} & \textit{-3.2} \\
    
            \hline
        \end{tabular}
        \label{tab:varying_conditions}
    \end{table}

\begin{figure} [t]
    \centering
    \includegraphics[scale=0.8]{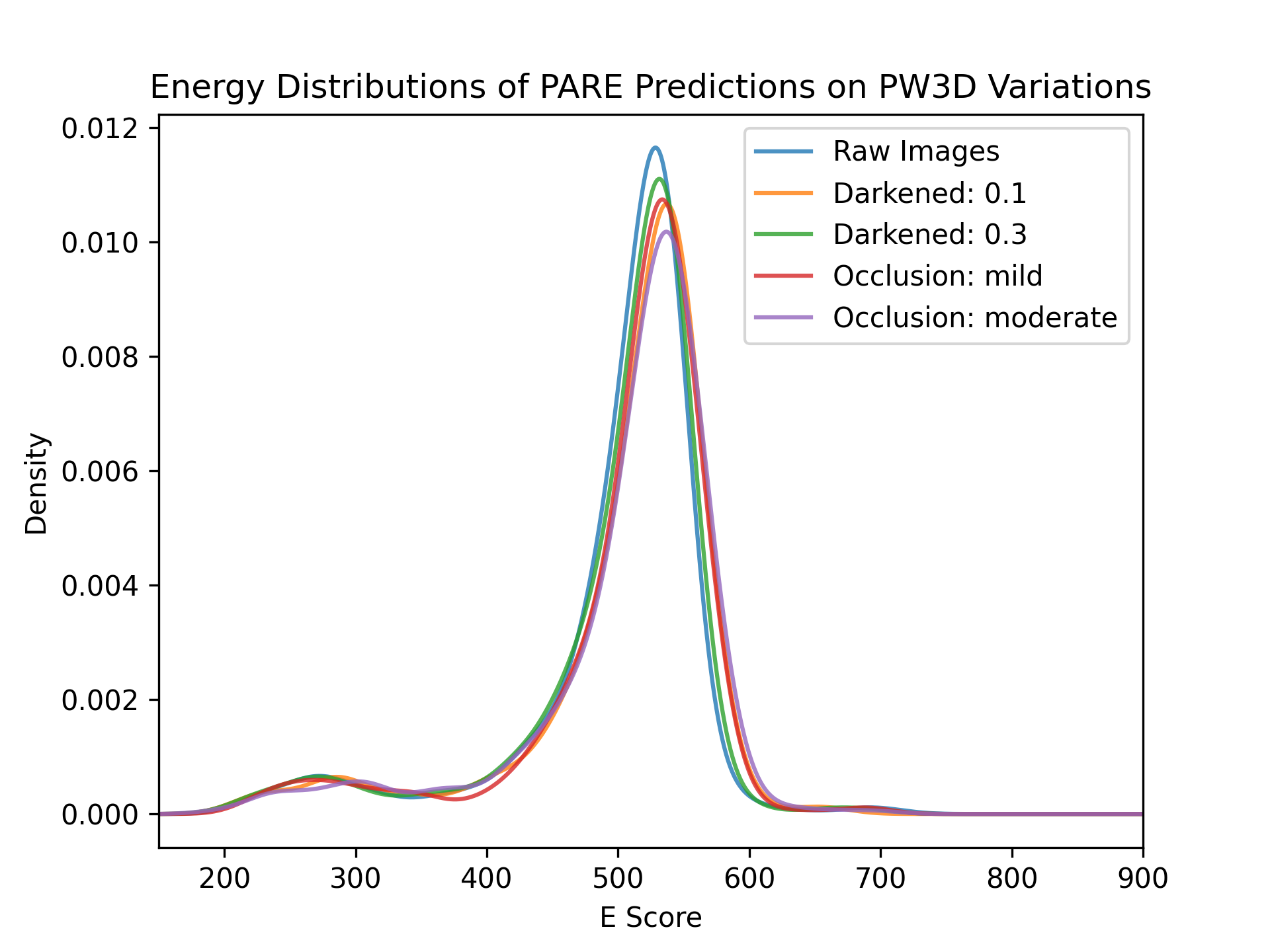}
    \caption{Visualization of the backbone estimation energy score PDF's for each of the condition variations added to the PW3D subset. The energy scores shift slightly but remain robust to the severe alterations to the raw images.}
    \label{fig:variations_E_densities}
\end{figure}

\textbf{Robustness of Energy Function to Varying Conditions.} The energy function is a central component in the proposed method and the consistency of its scores for similar poses under variations in environmental conditions is important. In order to understand the impact of such variations, we recorded the energy scores of the backbone predictions for the raw images and each of the variations tested on the 1000 sample PW3D subset described in the preceeding ablation study. We visualize the distributions via their PDF's in Fig. \ref{fig:variations_E_densities} and observe that the backbone prediction energy score distributions for the same set of images do not significantly shift under the drastic environmental changes. Further, the quantitative results in Tab. \ref{tab:varying_conditions} show that TTA continues to provide improvements in these scenarios.

\section{Conclusions and Future Work}

    This work has proposed a novel selective test-time adaptation framework for 3D human pose estimation. Our framework addresses the lack of 2D supervision available in practice by proposing a novel consistency supervision based on the predicted errors of distal and proximal keypoints. We also sidestep the prohibitively large computational overhead of existing TTA methods in two main ways. First, by consciously selecting only the difficult OOD samples for intensive adaptation and otherwise only performing a fast, forward-pass correction. Second, by using the backbone estimator in a frozen, black box manner and only tuning a lightweight external network when intensive adaptation is indeed performed. To separate OOD and ID data, we select samples according to the recently proposed energy function threshold. We show the correction network can be effectively trained with data already used for the backbone estimator training and yet still learn to significantly improve the estimations. Further, we demonstrate the effectiveness of the energy function in separating out hard samples and demonstrate that applying the proposed sample selection strategy to an existing TTA method massively improves its inference time while maintaining the majority of its improvement. Our experiments indicate that energy based thresholding could be effectively used to mitigate compute time increases of future adaptation methods, and could help target samples of interest. Finally, we quantitatively and qualitatively show that ESCAPE significantly improves the performance of several popular HPE models, outperforming previous methods on two public benchmark datasets.  However, there are improvements to be made in future works. ESCAPE only corrects the distal keypoints, while future works should propose methods which additionally correct errors in proximal keypoints. Moreover, one of the major limitations of the pre-trained HPE model is estimating the global orientation of 3D human pose. Future work should address this limitation while optimizing the network at the inference stage. In addition, a fixed energy function threshold was used across all backbones, but more robust or adaptive approaches to energy thresholding should be developed.

\section{CRediT Authorship Contribution Statement}

\textbf{Luke Bidulka:} Conceptualization, Methodology, Software, Investigation, Writing - original draft. \textbf{Mohsen Gholami:} Conceptualization, Methodology, Investigation, Writing - original draft. \textbf{Jiannan Zheng:} Investigation, Writing - review \& editing. \textbf{Martin McKeown:} Supervision, Resources. \textbf{Jane Wang:} Supervision, Writing - review \& editing.

\section{Acknowledgement}

This work was supported by the Natural Sciences and Engineering Research Council of Canada (NSERC grant RGPIN-2022-03049) and was partially funded by a grant from the Canadian Institute of Health Research (PI: MJM, CHRP 00035-007) and the John Nichol Chair in Parkinson's Research.

\bibliographystyle{elsarticle-num-names} 
\bibliography{elsarticle}

\end{document}